\documentclass[11pt]{article}

\usepackage[final]{acl}

\usepackage{times}
\usepackage{latexsym}
\usepackage[T1]{fontenc}
\usepackage[utf8]{inputenc}
\usepackage{microtype}
\usepackage{inconsolata}
\usepackage{graphicx}

\usepackage{pgfplots}
\usepgfplotslibrary{statistics}
\pgfplotsset{compat=1.18}

\usepackage[table]{xcolor}
\usepackage{booktabs}
\definecolor{s2lgreen}{RGB}{218,232,218}
\definecolor{s2lgreendark}{RGB}{205,222,205}
\usepackage{multirow}

\usepackage[most]{tcolorbox}
\usepackage{listings}
\usepackage{xcolor}

\usepackage{booktabs}
\usepackage{array}
\usepackage{amsmath}
\usepackage{makecell}
\usepackage{multirow}
\usepackage{tabularx}
\usepackage{xcolor}
\usepackage{float}

\usepackage{placeins}
\usepackage{dblfloatfix}
\title{Skill-to-LoRA: From Using Skills to Learning Behaviors for Token-Efficient LLM Agents}
\author{
  Tianyi Zhang$^{1}$\thanks{Equal contribution.},
  Zhonghao Qi$^{1}$\footnotemark[1] \\
  $^{1}$The Chinese University of Hong Kong \\
}

\newcommand{\skillmd}{\texttt{SKILL.md}}

\begin{document}
\maketitle

\begin{abstract}
Agent skills are commonly distributed as \texttt{SKILL.md} files: human-readable procedural documents that describe workflows, tools, resources, and domain conventions. While convenient for inspection and reuse, this design requires the same reusable procedure to be repeatedly injected into the runtime context. We propose \textit{Skill-to-LoRA} (S2L), a behavior-centric skill representation that replaces runtime skill text with skill-specific LoRA adapters. Rather than compressing the skill document itself, S2L models the behavioral change induced by the skill text: offline, the complete \texttt{SKILL.md} is used to synthesize skill-guided demonstrations; online, the full document is omitted and the corresponding LoRA adapter is dynamically loaded to activate the learned skill behavior.
We evaluate S2L with Qwen3.6-27B~ on a 21-skill subset of SWE-Skills-Bench. Compared with the no-skill and Full Skill Text baselines, S2L improves pass rate by 2.9 and 5.2 percentage points, respectively, while reducing per-step token cost by 6.6\% relative to Full Skill Text prompting. S2L matches or improves Full Skill Text on 18/21 skills and the no-skill baseline on 15/21 skills. Control experiments further show that the gains depend on skill-specific adapter alignment: Wrong-LoRA and Shared-LoRA both reduce performance. These results suggest that many procedural agent skills can be converted from runtime instructions into trainable, dynamically loadable behavioral modules. Code will be released upon acceptance.
\end{abstract}

\section{Introduction}

Recent agent systems increasingly package reusable procedural knowledge as skill libraries, where each skill combines natural-language guidance with optional scripts, resources, and domain conventions. In software-engineering agents, a skill is more than a hint to the language model. It can shape which files the agent inspects, how it edits a repository, which commands it runs, and when a verifier-facing artifact is complete~\citep{li2026skillsbenchbenchmarkingagentskills,han2026sweskillsbenchagentskillsactually}.

The usual deployment mechanism keeps the skill body in the prompt. When a task selects skill $s$, the agent receives the full \skillmd{} document $d_s$ in context, together with the repository state and tool interface. This choice is convenient and auditable, yet every inference step must reread the same reusable procedure. The cost extends beyond tokens. A long skill document can compete with local repository evidence, introduce mismatched templates, or widen the action space. SWE-Skills-Bench reports that even curated skill prompting is often neutral or harmful~\citep{han2026sweskillsbenchagentskillsactually}. Existing compression, retrieval, or caching strategies can reduce redundant context, but they still leave the core skill procedure as runtime text.

To address this problem, we propose S2L (Skill-to-LoRA), a behavior-centric skill representation method. Rather than directly compressing or summarizing skill text, S2L focuses on how the complete \skillmd{} changes the model's response behavior on the same task after being injected into the prompt. We treat this skill-induced behavioral change as a learnable functional bias and model it using lightweight LoRA parameters. During the offline stage, the system first uses the complete skill document to generate skill-guided behavioral demonstrations, and then distills these behavioral patterns into LoRA parameters through LoRA training.
During online inference, the agent keeps only lightweight skill metadata, while the full procedural text is removed from the runtime prompt. Instead, the corresponding LoRA adapter is dynamically loaded to activate the associated skill behavior. In this way, skill invocation shifts from repeatedly injecting textual instructions into the prompt to activating parameterized behavioral representations. On SWE-Skills-Bench, S2L improves the pass rate by 5.2 percentage points over the Full Skill Text baseline while reducing per-step token consumption by 6.6\%. These results suggest that parameterized skill representations can not only reduce context overhead, but may also mitigate interference caused by long skill documents during model inference.

The contributions of this paper are:
\begin{itemize}
    \item We introduce S2L, a behavior-centric Skill-to-LoRA approach that represents a skill by the behavioral change it induces in the model, rather than by its natural-language text alone.
    \item We propose a skill-based self-distillation pipeline that automatically generates skill-guided behavioral demonstrations from \texttt{SKILL.md} without large-scale human annotation.
    \item We convert each skill into a lightweight skill-specific LoRA adapter, shifting runtime skill invocation from repeated prompt injection to dynamic adapter activation.
    \item We evaluate S2L on SWE-Skills-Bench and show that it improves pass rate over Full Skill Text prompting while reducing per-step token consumption.
\end{itemize}
\section{Background and Related Work}

\begin{figure*}[!tbp]
\centering
\includegraphics[width=1\linewidth]{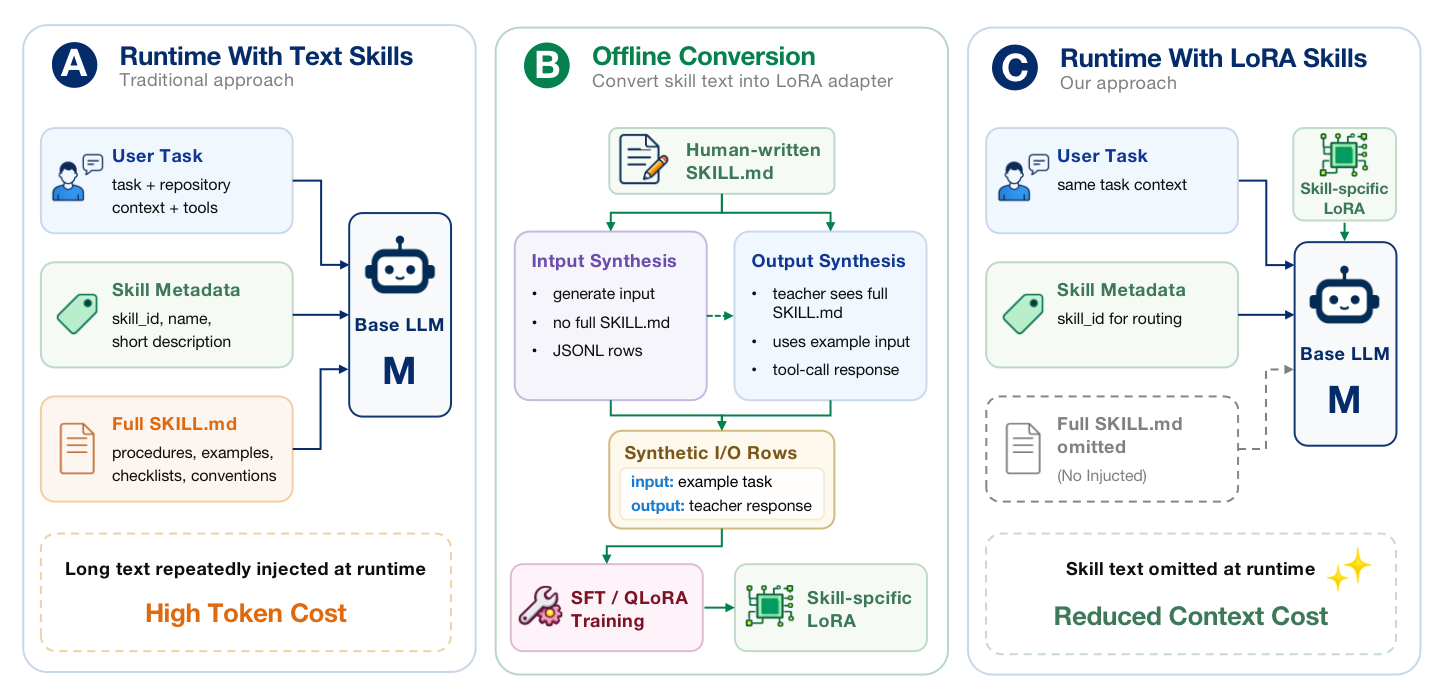}
\caption{\textbf{Text skill and LoRA skill representations.} Text skills keep the prompt, metadata, and full skill body in every agent step. LoRA skills use the skill body and demonstrations offline, then keep only prompt/metadata at runtime while loading LoRA weights onto the model.}
\label{fig:text-vs-lora}
\end{figure*}

\paragraph{Agent systems and skill definitions.}
We follow the file-based notion of skills used by Anthropic and by
SWE-Skills-Bench. A skill is a directory centered on a required \skillmd{} file
with YAML metadata and procedural instructions, plus optional scripts,
templates, and reference resources that can be loaded when relevant
\citep{anthropic2026claudeskills,han2026sweskillsbenchagentskillsactually}.
In software-engineering tasks, this package view is operational: a selected skill can affect where an agent searches, how it
edits a repository, which commands it runs, and how it validates a patch.
SWE-agent further shows that the agent-computer interface can shape repository
navigation, editing, and testing behavior~\citep{yang2024sweagent}. We study
this file-based skill interface while replacing only the repeated
natural-language body of the selected \skillmd{}.

\paragraph{Skill benchmarks.}
SkillsBench and SWE-Skills-Bench evaluate whether procedural skill injection
improves executable agent tasks
\citep{li2026skillsbenchbenchmarkingagentskills,han2026sweskillsbenchagentskillsactually}.
SkillsBench spans multiple domains and studies how curated skills, self-generated skills, and skill granularity affect task success. SWE-Skills-Bench is more directly aligned with our setting: it evaluates software-engineering
tasks paired with curated skill documents under controlled with-skill and no-skill conditions, building on the repository-level task setting introduced by SWE-bench~\citep{jimenez2024swebench}. The single-skill assignment in SWE-Skills-Bench is central to our design,
because one selected \skillmd{} body can be replaced by one skill-specific LoRA
module without introducing multi-skill composition. SWE-Skills-Bench also motivates the replacement question directly: it reports limited average gains,
many skills with no improvement, and substantial token overhead for full skill-text prompting.

\paragraph{Experience-derived textual skills and skill memories.}
Agents can also acquire reusable procedural knowledge from interaction experience. Trace2Skill distills execution experience into transferable declarative skill directories, and SkillNet studies infrastructure for creating, evaluating, and connecting skills
\citep{ni2026trace2skilldistilltrajectorylocallessons,liang2026skillnetcreateevaluateconnect}.
Skill set optimization extracts high-reward subtrajectories into subgoal-level
in-context instructions, while SkillRL and Skill1 train agents to evolve skill
use and skill memories under reinforcement learning
\citep{nottingham2024skillsetoptimization,xia2026skillrlevolvingagentsrecursive,shi2026skill1}.
SkillLearnBench further evaluates continual skill generation and acquisition
from agent experience~\citep{zhong2026skilllearnbenchbenchmarkingcontinuallearning}.
These works are related because they study reusable procedural knowledge, but
their skill objects are experience-derived textual instructions, memories, or
skill banks rather than the curated file-based SWE-Skills \skillmd{} bodies that
S2L replaces. Instead of learning from benchmark rollouts or verifier signals,
S2L uses the original skill markdown to synthesize workflow demonstrations for
LoRA training.

\paragraph{Parameter-side context and skill internalization.}
LoRA and QLoRA make skill-level adaptation practical by training small
low-rank updates on top of a frozen backbone
\citep{hu2021loralowrankadaptationlarge,dettmers2023qloraefficientfinetuningquantized}.
A growing body of work studies how recurring prompts, task descriptions,
documents, or long contexts can be moved from runtime text into model
parameters. PromptIntern internalizes recurring prompts during fine-tuning;
Text-to-LoRA and Doc-to-LoRA map task descriptions or documents to LoRA
parameters; and deep context distillation trains reusable parameter-side
knowledge modules for long contexts
\citep{zou-etal-2024-promptintern,charakorn2025texttolorainstanttransformeradaption,charakorn2026doctoloralearninginstantlyinternalize,caccia2025trainingplugnplayknowledgemodules}.
SKILL0 is closest in motivation: it uses an agentic reinforcement-learning
curriculum that gradually withdraws skill context and trains the agent from
environment feedback~\citep{lu2026skill0incontextagenticreinforcement}. S2L uses
a supervised route instead. It trains skill-specific LoRA weights from
skill-conditioned synthetic workflow demonstrations, and at inference time the
benchmark-selected skill identity activates the matched weights while the full
\skillmd{} body is omitted.
\section{Skill-to-LoRA}

Skill-to-LoRA (S2L) converts natural-language skill instructions into dynamically loadable parameterized representations, allowing the model to acquire skill-specific behavior without repeatedly injecting the full skill document during inference, thereby reducing runtime token overhead. Figure~\ref{fig:text-vs-lora} contrasts the traditional runtime text-skill paradigm (Figure~\ref{fig:text-vs-lora}A) with the proposed S2L pipeline, which consists of offline Skill-to-LoRA conversion and runtime LoRA-based skill activation (Figure~\ref{fig:text-vs-lora}B-C). In the conventional Full Skill Text setting, the complete \texttt{SKILL.md} is concatenated into the prompt for every model call, which increases context length and becomes increasingly expensive as skill documents and skill libraries grow. To address this problem, S2L transfers the procedural behavior induced by each skill document into a lightweight LoRA module, enabling the model to reproduce the corresponding behavior without directly accessing the full skill text.
The overall framework consists of two stages: offline training and online inference. During offline training, we propose \textit{skill-based self-distillation}. The system first uses the complete \texttt{SKILL.md} to automatically generate task-level inputs and target outputs, allowing the base model to exhibit the behavior induced by Full Skill Text prompting. The base model is then frozen, and only the LoRA parameters are trained to learn and store these procedural behavioral biases. During online inference, the system no longer injects the complete skill document into the prompt. Instead, it dynamically loads the corresponding LoRA adapter according to the target skill, enabling skill-conditioned task execution with substantially shorter runtime context.

\subsection{Problem Formulation}
In LLM-based agent systems, a skill is typically represented as a natural-language document that specifies the workflow, tool-use patterns, configuration conventions, and verification steps required for a particular class of tasks. In this work, we denote the set of skills available to the agent as $\mathcal{S}=\{s_1,s_2,\dots,s_N\}$, where each skill $s_i$ is associated with a \texttt{SKILL.md} document $t_i$. The set of all skill documents is denoted as $\mathcal{T}=\{t_1,t_2,\dots,t_N\}$.

Traditional Full Skill Text prompting directly concatenates the complete skill document into the runtime prompt:
\[
y = M(x \oplus t_i),
\]
where $M$ denotes the base language model, $x$ denotes the task input, $\oplus$ denotes prompt concatenation, and $y$ denotes the model output. This work studies how to preserve the procedural task-solving behavior induced by the skill document $t_i$ without explicitly accessing the complete skill text at runtime, while reducing context length and token overhead.

\subsection{Skill-Based Self-Distillation}

The goal of S2L is to convert the behavior induced by Full Skill Text prompting into a loadable parameterized representation, allowing the model to reproduce the corresponding task-solving behavior without repeatedly reading the complete skill document at inference time. We formulate this process as a supervised learning problem and, inspired by knowledge and feature distillation methods~\citep{hinton2014distilling,romero2015fitnets}, propose \textit{skill-based self-distillation}. In this framework, an LLM conditioned on the full skill document serves as the teacher model, while the frozen base model equipped with trainable LoRA parameters serves as the student model. The student learns to reproduce the procedural behavioral patterns exhibited by the teacher using only the task input.

\subsubsection{Synthetic Data Generator}

Training data consists of two parts: an input and a label. The input is a task request that is expected to require a specific skill, while the label is the response generated by the model after reading both the task and the complete skill document. The entire data generation process is fully automatic and does not require manual annotation. To construct such data, we design two LLM-based data generation agents, one for generating training inputs and the other for generating the corresponding labels.
The first agent generates task inputs. Given the complete \texttt{SKILL.md} of a skill, it constructs a set of task requests that naturally require the target skill. The prompt template used by this agent is shown below:

\begin{tcolorbox}[
  enhanced,
  colback=green!1,
  colframe=green!20!black,
  boxrule=0.3pt,
  arc=2mm,
  left=1mm,
  right=1mm,
  top=0.8mm,
  bottom=0.8mm,
  boxsep=1mm,
  title=\textbf{Training Input Generation Prompt},
  fonttitle=\small\bfseries,
  coltitle=black,
  colbacktitle=green!6,
  breakable
]

\footnotesize

\begin{lstlisting}[
  basicstyle=\scriptsize\ttfamily,
  breaklines=true,
  columns=fullflexible,
  keepspaces=true
]
Generate {n} diverse example inputs that would naturally trigger this skill.

Rules:
- Use only the skill document.
- Do not use benchmark task instructions.
- Do not include the full skill text in example_input.
- Make the examples realistic user/task requests for applying this reusable skill.
- Vary repository context, failure modes, and desired artifact style.

skill_id: {skill_id}

SKILL.md: {skill_text}
\end{lstlisting}

\end{tcolorbox}

The second agent generates training labels. It takes the task queries generated by the first agent and the complete \texttt{SKILL.md} as input, and asks the LLM to produce target responses according to the workflow and requirements specified by the skill. These responses serve as behavioral demonstrations of how the skill should be applied in concrete task scenarios. The prompt template used by this agent is shown below:

\begin{tcolorbox}[
  enhanced,
  colback=green!1,
  colframe=green!20!black,
  boxrule=0.3pt,
  arc=2mm,
  left=1mm,
  right=1mm,
  top=0.8mm,
  bottom=0.8mm,
  boxsep=1mm,
  title=\textbf{Training Label Generation Prompt},
  fonttitle=\small\bfseries,
  coltitle=black,
  colbacktitle=green!6,
  breakable
]

\footnotesize

\begin{lstlisting}[
  basicstyle=\scriptsize\ttfamily,
  breaklines=true,
  columns=fullflexible,
  keepspaces=true
]
Use the full skill document to respond to each example input.

The student LoRA will later see only the short skill id plus the example input, not the SKILL.md.
Your responses should demonstrate the concrete behavior induced by the skill text: when to apply it, what steps to take, what files/tools/tests to inspect, and how to verify the result.
Keep each output concise enough for SFT: use Markdown steps and short snippets only, under 700 words.

skill_id: {skill_id}

SKILL.md: {skill_text}

Example inputs: {example_inputs}
\end{lstlisting}

\end{tcolorbox}

Finally, each training example consists of one task input and its corresponding target output, which are then used for subsequent LoRA training.

\subsubsection{LoRA Behavioral Distillation}
During training, we freeze the base model parameters and update only the lightweight LoRA parameters. The model receives only the task input $x_i$, and the objective is to make its output approximate the teacher-generated target output $y_i$. This process can be written as:
\[
\hat{y}_i = (M + A_s)(x_i),
\]
where $A_s$ denotes the LoRA adapter associated with skill $s$, $(M + A_s)$ denotes the student model obtained by loading the adapter onto the frozen base model $M$, and $\hat{y}_i$ denotes the output generated from the task input $x_i$.

We optimize a standard causal language-modeling objective on the assistant completion tokens only. For each training pair $(x_i, y_i)$, the prompt tokens from $x_i$ are masked out, and the loss is computed only over the target output $y_i$:
\[
\mathcal{L}_s
=
-\sum_i \sum_{t=1}^{|y_i|}
\log P\!\left(y_{i,t}\mid x_i, y_{i,<t}; M + A_s\right).
\]
By minimizing this loss, the LoRA adapter gradually absorbs the behavioral bias induced by Full Skill Text prompting. After training, the procedural knowledge of the skill is no longer represented as long natural-language instructions in the runtime prompt, but is instead stored in the corresponding lightweight LoRA parameters. For all skills in the skill library, we train one independent LoRA adapter for each skill, resulting in an adapter library
\[
\mathcal{A}=\{A_s \mid s \in \mathcal{S}\},
\]
where each adapter stores the procedural behavior associated with its corresponding skill and can be dynamically loaded during runtime execution.

\subsection{Skill-Conditioned LoRA Activation}

At inference time, the system uses the skill id associated with the current task to retrieve the corresponding LoRA adapter $A_s$ from the adapter library $\mathcal{A}$, and dynamically loads it onto the frozen base model $M$. The adapter $A_s$ encodes the procedural behavioral bias associated with skill $s$, including workflow patterns, tool-use preferences, file-inspection strategies, and verification behavior. Once activated, the adapter steers the base model toward the behavior induced by the corresponding skill, without requiring access to the full \texttt{SKILL.md} document.
The runtime prompt therefore keeps only lightweight skill metadata for routing, while the procedural knowledge is supplied by the selected LoRA adapter. Since all skills share the same base model $M$, the system can switch between skills by dynamically loading different lightweight adapters, reducing runtime context length and token cost.

\begin{table*}[tbp]
\centering
\small
\setlength{\tabcolsep}{3.5pt}
\renewcommand{\arraystretch}{1.04}
\begin{tabular*}{\textwidth}{@{\extracolsep{\fill}} r l c c c r r r r @{}}
\toprule
\multirow{2}{*}{\#} & \multirow{2}{*}{Skills}
& \multicolumn{3}{c}{Pass $\uparrow$}
& \multicolumn{2}{c}{$\rho$ (\%) $\downarrow$}
& \multicolumn{2}{c}{CNG $\uparrow$} \\
\cmidrule(lr){3-5}
\cmidrule(lr){6-7}
\cmidrule(lr){8-9}
& & V & T & S2L & T & S2L & T & S2L \\
\midrule
1  & \texttt{risk-metrics-calculation}      & 5/10 & 1/10 & 5/10 & 72.10 & 9.12   & -0.55 & 0.00 \\
2  & \texttt{gitlab-ci-patterns}            & 5/10 & 6/10 & 7/10 & 12.53 & 41.09  & 0.80  & 0.49 \\
3  & \texttt{prompt-engineering-patterns}   & 3/10 & 3/10 & 3/10 & 28.38 & 40.31  & 0.00  & 0.00 \\
4  & \texttt{similarity-search-patterns}    & 0/10 & 0/10 & 0/10 & 32.09 & 0.28   & 0.00  & 0.00 \\
5  & \texttt{distributed-tracing}           & 2/10 & 2/10 & 4/10 & -38.59 & 24.45 & 0.00  & 0.82 \\
6  & \texttt{tdd-workflow}                  & 1/10 & 1/10  & 4/10 & 11.11 & 5.05  & 0.00  & 5.94 \\
7  & \texttt{istio-traffic-management}      & 6/10 & 8/10 & 7/10 & -3.06  & 9.91  & 6.53  & 1.01 \\
8  & \texttt{service-mesh-observability}    & 0/10 & 0/10 & 0/10 & 47.26 & 24.60  & 0.00  & 0.00 \\
9  & \texttt{python-background-jobs}        & 7/10 & 5/10 & 5/10 & -14.20 & -18.08 & -1.41 & -1.11 \\
10 & \texttt{turborepo}                     & 2/10 & 1/10 & 3/10 & 58.66 & -12.72 & -0.17 & 0.79 \\
11 & \texttt{python-observability}          & 7/10 & 3/10 & 3/10 & 18.51 & 1.78   & -2.16 & -22.51 \\
12 & \texttt{fix}                           & 1/10 & 1/10 & 2/10 & -21.16 & -24.87 & 0.00  & 0.40 \\
13 & \texttt{bash-defensive-patterns}       & 5/10 & 3/10 & 4/10 & 12.95 & -26.60 & -1.54 & -0.38 \\
14 & \texttt{gitops-workflow}               & 0/10 & 0/10 & 0/10 & -3.50  & -40.67 & 0.00  & 0.00 \\
15 & \texttt{python-resilience}             & 3/10 & 3/10 & 2/10 & -14.67 & -34.30 & 0.00  & -0.29 \\
16 & \texttt{bazel-build-optimization}      & 1/10 & 0/10 & 0/10 & -3.91  & -24.76 & -2.56 & -0.40 \\
17 & \texttt{v3-performance-optimization}   & 0/10 & 0/10 & 0/10 & -33.27 & -45.91 & 0.00  & 0.00 \\
18 & \texttt{changelog-automation}          & 2/10 & 6/10 & 6/10 & 7.34  & -1.32  & 5.45  & 30.19 \\
19 & \texttt{implementing-agent-modes}      & 2/10 & 3/10 & 3/10 & 88.53 & 21.91  & 0.11  & 0.46 \\
20 & \texttt{python-anti-patterns}          & 3/10 & 4/10 & 1/10 & 8.50  & -49.41 & 1.18  & -0.40 \\
21 & \texttt{slo-implementation}            & 4/10 & 4/10 & 6/10 & 15.66 & -2.51  & 0.00  & 7.96 \\
\midrule
\multicolumn{2}{r}{\textbf{Aggregate}}
& \textbf{59/210} & \textbf{54/210} & \textbf{65/210}
& \textbf{13.39} & \textbf{-4.89}
& \textbf{-0.18} & \textbf{0.58} \\
\bottomrule
\end{tabular*}
\caption{
Pass and token-cost comparison across 21 SWE-Skills-Bench skills.
V denotes the Vanilla LLM baseline, T denotes the Full Skill Text setting, and S2L denotes our method.
Pass reports the number of solved tasks for each skill.
$\rho$ reports the relative token-cost change compared with V, where lower values indicate lower token cost.
CNG denotes cost-normalized gain; positive values indicate pass-rate improvement relative to V, while negative values indicate pass-rate degradation.
}
\label{tab:pass_cost_comparison}
\end{table*}

\section{Experiments Setup}

\paragraph{Benchmark.}
We evaluate S2L on a fixed 21-skill subset from SWE-Skills-Bench, covering 210 tasks in total. The subset includes all 7 skills that show positive pass-rate gains under skill prompting, plus 14 zero-delta skills selected by token-cost extremes: 7 with the largest cost increases and 7 with the largest cost decreases. This design preserves the benchmark observation that correctness gains and token overhead are often decoupled.

\paragraph{Compared Methods.}
We compare three main conditions: \textbf{Vanilla LLM}, which uses no skill; \textbf{Full Skill Text}, which injects the complete \texttt{SKILL.md}; and \textbf{S2L}, which replaces the skill body with a skill-specific LoRA adapter. All conditions share the same task files, mounted resources, OpenCode interface, verifier, context budget, and output budget. We also include two controls: \textbf{Wrong-LoRA}, which loads the least-similar skill adapter, and \textbf{Shared-LoRA}, which trains one adapter over pooled skill data.

\paragraph{Base Model and Runtime Environment.}
All experiments use Qwen3.6-27B~\citep{qwen2026qwen36} as the frozen base model and are executed through OpenCode with a vLLM OpenAI-compatible server~\citep{kwon2023efficient}. The runtime configuration uses a 32K OpenCode context budget, a 4K output budget, and a 36K server-side maximum sequence length, with Qwen thinking mode disabled. These settings are fixed across all compared conditions. Each task is first executed under a 3600-second wall-clock budget. Tasks that fail to produce a final verifier report within this limit are rerun with a 7200-second budget. Remaining timeout or missing-report cases are counted as failures in the final evaluation tables.

\paragraph{Training Configuration.}
All LoRA modules are trained with QLoRA. Each synthesized example is formatted as a Qwen chat-style training instance, where the generated task input serves as the user message and the generated target output serves as the assistant response. During training, loss is computed only on the assistant completion tokens. For each skill, we train an independent LoRA adapter with rank $r=16$, $\alpha=32$, dropout 0, and target modules \texttt{q\_proj} and \texttt{v\_proj}. We use a maximum sequence length of 4096, learning rate $10^{-4}$, batch size 1, gradient accumulation 16, one training epoch, and paged AdamW 8-bit optimization. Each trained adapter contains approximately 6.03M trainable parameters, corresponding to about 0.022\% of the base model, and occupies roughly 24MB of storage.

\paragraph{Runtime Serving.}
During inference, the frozen base model is served through vLLM~\citep{kwon2023efficient} with dynamic LoRA loading enabled. For each task, the selected skill id determines which LoRA adapter is retrieved from the adapter library and activated during execution. Unlike Full Skill Text prompting, S2L does not inject the complete \texttt{SKILL.md} into the runtime prompt; only lightweight skill metadata is retained for routing, while procedural behavior is supplied through the activated LoRA adapter. Reported runtime token statistics include only inference-time token usage after offline synthesis and LoRA training have been completed.

\paragraph{Metrics.}
We report pass rate and runtime token cost following SWE-Skills-Bench. Pass rate is a binary task-level metric based on the deterministic verifier, and token cost $C$ measures the average per-step input and output tokens consumed during inference. We use relative token change $\rho$ to indicate the percentage change in token cost relative to Vanilla LLM. To relate performance gain to token-cost change, we introduce Cost-Normalized Gain (CNG):
\[
\mathrm{CNG}=\frac{P_{\mathrm{cond}}-P_{\mathrm{base}}}{|\rho|/100},
\]
where $P_{\mathrm{base}}$ and $P_{\mathrm{cond}}$ are the pass rates of Vanilla LLM and the evaluated condition. Higher CNG indicates larger pass-rate improvement per unit of relative token-cost change.

\section{Results and Analysis}

\subsection{Main Results}
\begin{figure}
    \centering
    \includegraphics[width=1\linewidth]{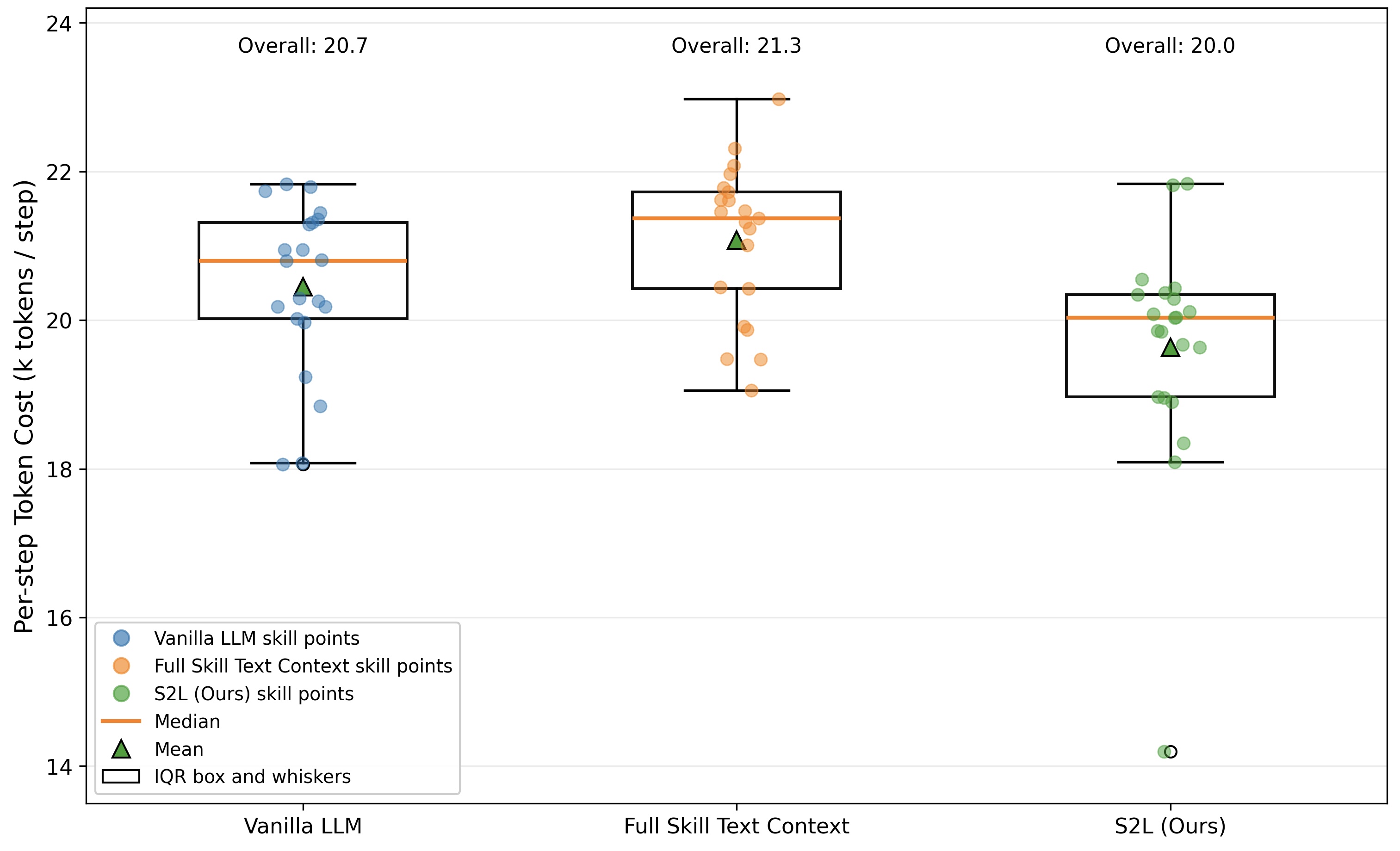}
    \caption{Per-skill per-step token cost distribution. S2L achieves the lowest mean and median token cost per execution step.}
    \label{fig2}
\end{figure}

Table~\ref{tab:pass_cost_comparison} reports the overall evaluation results across 21 SWE-Skills-Bench skills. S2L achieves the highest aggregate pass rate, solving $65/210$ tasks, compared with $59/210$ for Vanilla LLM and $54/210$ for Full Skill Text prompting. This result suggests that procedural skill behavior can be effectively transferred from long natural-language skill documents into lightweight LoRA adapters, and that the resulting parameterized skill representation can match or exceed the original prompt-based skill conditioning.
S2L also demonstrates substantially better runtime efficiency. Relative to Vanilla LLM, Full Skill Text prompting increases runtime token cost by $13.39\%$, whereas S2L reduces token cost by $4.89\%$. This trend is further reflected in the aggregate CNG score: S2L achieves a positive CNG of $0.58$, while Full Skill Text obtains $-0.18$. These results indicate that S2L achieves larger pass-rate gains per unit of token-cost change, yielding a more favorable trade-off between task performance and runtime efficiency.
At the skill level, S2L shows particularly strong improvements on workflow-heavy skills such as \texttt{tdd-workflow}, \texttt{distributed-tracing}, and \texttt{slo-implementation}. In several cases, S2L improves task success rate while simultaneously reducing runtime token cost, suggesting that the learned adapters capture reusable procedural behavior rather than merely reproducing prompt context.
Figure~\ref{fig2} further shows the distribution of per-skill per-step token cost across methods. Compared with Full Skill Text prompting, S2L shifts the overall token-cost distribution toward lower values while maintaining comparable skill-level variance. Full Skill Text has a higher median token cost and a more pronounced upper tail, reflecting the overhead introduced by repeatedly injecting long skill documents into the runtime context. In contrast, S2L achieves lower average token cost through parameterized skill activation, without repeated text injection at inference time.

\subsection{Analysis}

\begin{figure}
    \centering
    \includegraphics[width=1\linewidth]{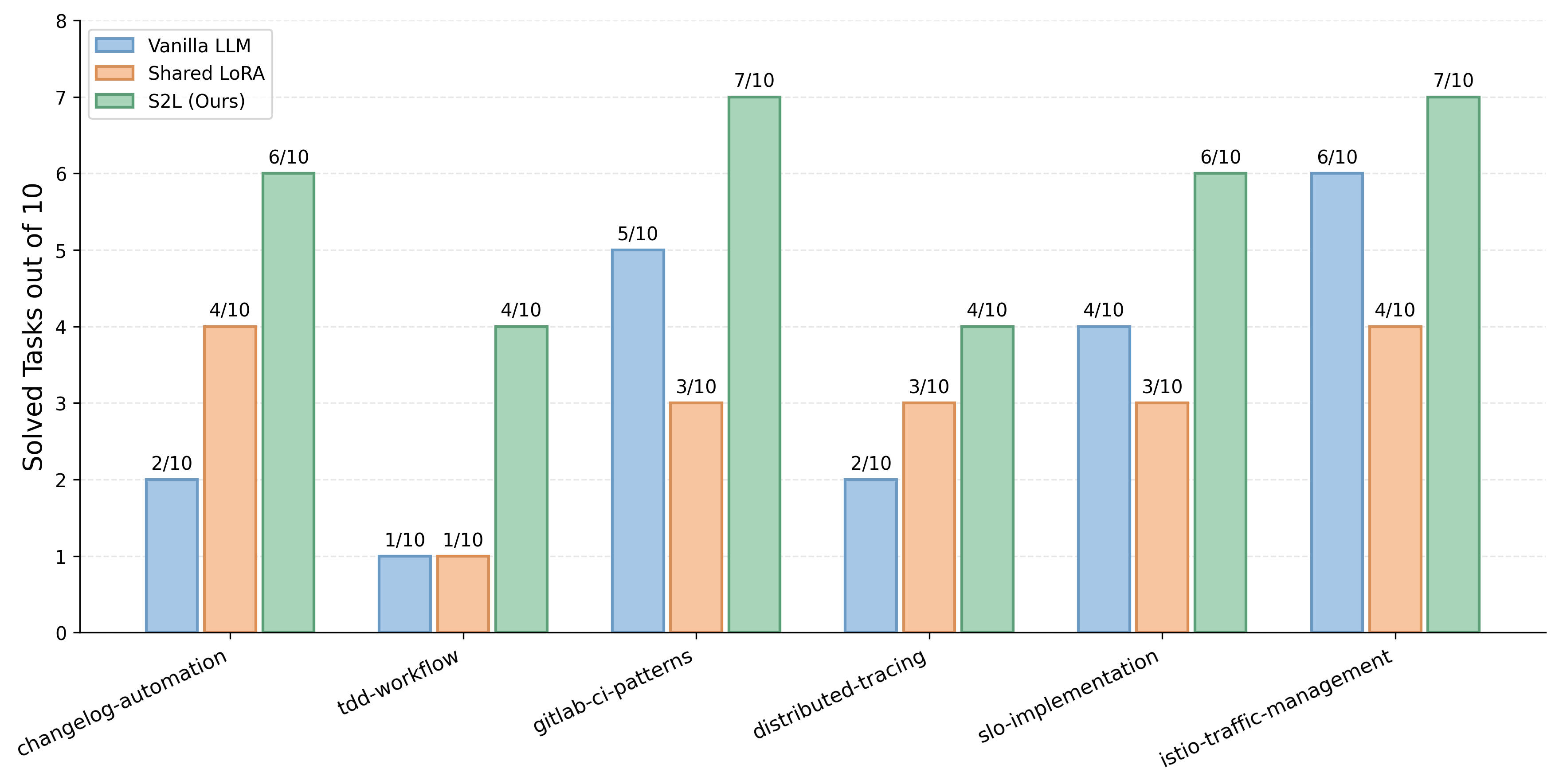}
    \caption{Comparison between Shared-LoRA and skill-specific S2L adapters on six representative skills.}
    \label{fig:shared-lora}
\end{figure}

\paragraph{Effect of Skill-Specific Adapters}
Figure~\ref{fig:shared-lora} compares S2L with a Shared-LoRA baseline trained on pooled demonstrations from multiple skills. Across all six control skills, S2L consistently outperforms the shared adapter setting. For example, \texttt{gitlab-ci-patterns} improves from $3/10$ under Shared-LoRA to $7/10$ under S2L, while \texttt{distributed-tracing} improves from $3/10$ to $4/10$. Similar trends appear for \texttt{changelog-automation}, \texttt{tdd-workflow}, and \texttt{slo-implementation}.
These results suggest that the observed gains do not come from generic SWE adaptation or simply loading additional LoRA parameters. Instead, procedural skill behavior appears to be highly skill-specific. Skills such as CI configuration, tracing, changelog generation, and SLO implementation each require distinct workflow structures, file-edit patterns, and verification procedures that are difficult to compress into a single shared adapter. Training independent adapters therefore provides a more effective parameterization of procedural behavior than learning a unified representation across heterogeneous skills.

\begin{figure}
    \centering
    \includegraphics[width=1\linewidth]{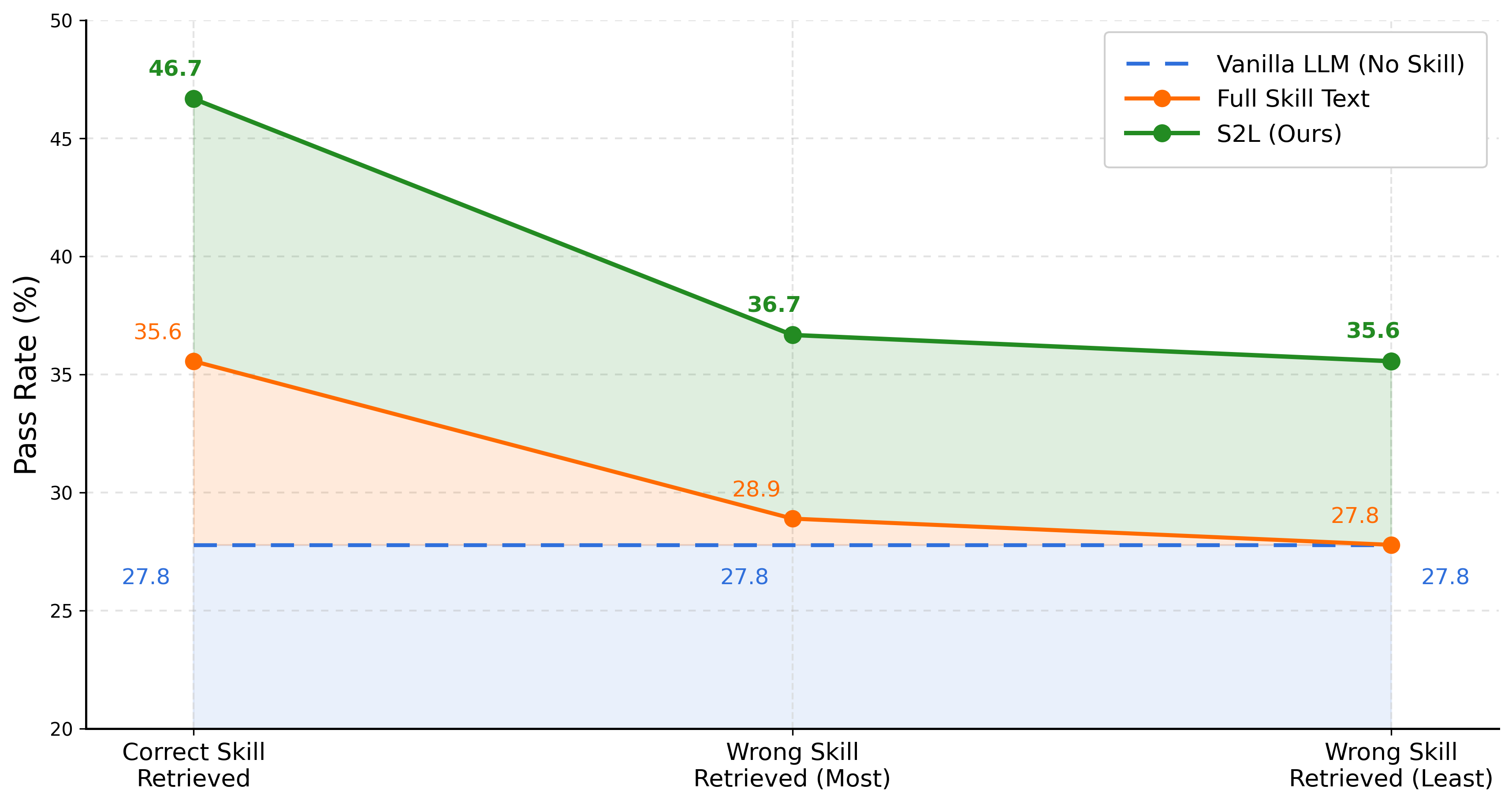}
    \caption{Robustness under retrieval mismatch. Most/Least denote the most-/least-similar wrong skills; S2L remains well above Vanilla LLM while Full Skill Text collapses toward the baseline.}
    \label{fig:retrieval-mismatch}
\end{figure}

\paragraph{Robustness under Retrieval Mismatch}
Figure~\ref{fig:retrieval-mismatch} evaluates robustness when the retrieved skill does not match the target task. We compare retrieval of the correct skill, the most similar wrong skill, and the least similar wrong skill. Under retrieval mismatch, Full Skill Text prompting rapidly collapses toward the Vanilla LLM baseline, decreasing from $35.6\%$ to $28.9\%$ and eventually to $27.8\%$. In contrast, S2L maintains substantially higher pass rates, decreasing from $46.7\%$ to $36.7\%$ and remaining at $35.6\%$ even under the least-similar wrong-skill condition.
This result suggests that parameterized skill behavior is more robust to imperfect skill routing than direct textual prompting. Full Skill Text prompting depends heavily on exact prompt-task alignment; once the injected skill text becomes mismatched, the additional context provides little useful procedural guidance and may even introduce irrelevant behavioral bias. In contrast, LoRA-based skill activation appears to encode higher-level procedural regularities that partially transfer across related tasks, allowing S2L to retain substantial capability even when the retrieved skill is incorrect.

\subsubsection{Ablation Study}

Table~\ref{tab:ablation} ablates LoRA rank and synthetic data scale on \texttt{gitlab-ci-patterns}. The results show that procedural skill behavior can be effectively parameterized with compact adapters and does not require excessive synthetic data. With 64 training rows, increasing the LoRA rank from 8 to 16 slightly improves the score from 93.2 to 93.5, while rank 32 brings no further gain, suggesting that a small adapter already captures most workflow and verification patterns. Increasing data scale also does not yield monotonic improvement: 16-row and 32-row settings underperform the selected 64-row configuration, and expanding to 128 rows reduces the score to 88.2. This suggests that the procedural behavior space of a skill may be limited; beyond a certain point, additional demonstrations are more likely to repeat existing workflows, artifact schemas, or verification patterns than introduce new behavioral structure, leading to behavioral distribution imbalance. Overall, the 64-row, rank-16 setting provides the best trade-off between parameter efficiency and performance, matching the best higher-rank score with only about 6.03M trainable parameters.
\begin{table}[t]
\centering
\small
\setlength{\tabcolsep}{7pt}
\renewcommand{\arraystretch}{1.08}
\begin{tabular}{lrrrc}
\toprule
Configuration & Rows & Rank & Params & Score \\
\midrule
Small data     & 16  & 16 & 6.03M  & 88.2 \\
Medium data    & 32  & 16 & 6.03M  & 86.7 \\
Low-rank       & 64  & 8  & 3.02M  & 93.2 \\
\rowcolor{s2lgreen}
\textbf{Selected} & \textbf{64} & \textbf{16} & \textbf{6.03M} & \textbf{93.5} \\
High-rank      & 64  & 32 & 12.06M & 93.5 \\
Large data     & 128 & 16 & 6.03M  & 88.2 \\
\bottomrule
\end{tabular}
\caption{
Compact LoRA configuration sweep on \texttt{gitlab-ci-patterns}.
The selected 64-row, rank-16 setting matches the best score while using fewer parameters than the high-rank alternative.
}
\label{tab:ablation}
\end{table}

\FloatBarrier
\section{Discussion}

The results of S2L suggest that skills do not necessarily need to remain as natural-language text inside the runtime prompt. Rather than repeatedly injecting the complete \texttt{SKILL.md} into context, S2L focuses on the behavioral change induced by the skill text and converts this change into a learnable parameterized representation. Our experiments show that, even after removing the full skill document, the model can still reproduce similar workflows, tool-use patterns, and verification strategies through dynamic LoRA adapter activation alone. This indicates that many procedural skills can be interpreted as stable behavioral priors instead of instructions that must be reread during inference. The advantages of S2L are most visible in tasks with stable workflows, explicit artifact schemas, and clear verification patterns, where a compact action prior can be effectively distilled from a small number of demonstrations into lightweight LoRA parameters. At the same time, existing human-written skills also emerge as a practical source of self-supervision. Skill documents already contain sufficient workflow and verification structure to automatically synthesize behavioral demonstrations without large-scale manual annotation. This suggests that future skill libraries may evolve beyond prompt libraries into trainable, loadable, and composable repositories of parameterized behaviors. S2L also changes the system-level management of skills. In traditional text-based skill prompting, disabling a skill is fundamentally a context-cleaning problem; in S2L, the skill no longer persists inside the runtime prompt and is instead dynamically activated as an adapter, making skill management closer to adapter routing than context management. As scalable multi-LoRA serving systems continue to improve~\citep{sheng2023slora,chen2023punica}, parameterized skill representations may not only reduce token overhead, but also provide a more modular foundation for deploying and composing large-scale agent skill ecosystems.
\FloatBarrier

\section{Conclusion}

This paper presents S2L (Skill-to-LoRA), a behavior-centric approach that converts skill-induced model behavior into lightweight LoRA adapters. Instead of repeatedly injecting complete \texttt{SKILL.md} documents at inference time, S2L uses skill-based self-distillation to synthesize behavioral demonstrations and store the resulting workflows, tool-use patterns, and verification strategies in parameters. Experiments on SWE-Skills-Bench show that S2L improves pass rate while reducing per-step token consumption compared with Full Skill Text prompting. These results suggest that many procedural skills can move beyond runtime text prompts.

\section*{Limitations}

S2L is most effective for procedural skills with relatively stable workflows. In such tasks, skills usually induce consistent artifact schemas, tool-use patterns, and verification structures, making them easier to compress into stable parameterized behavioral representations. In contrast, skills that rely heavily on concrete code examples, flexible syntax transfer, or open-ended reasoning may still benefit from runtime text, which can provide richer local context and example retrieval capabilities. Moreover, distilling skill behavior into LoRA parameters is fundamentally a form of behavioral compression. The adapter learns the dominant behavioral effect induced by the skill rather than preserving every detail of the original skill document, so rare edge cases or highly specific configuration patterns may not be fully retained. Parameterized skill representations also reduce direct interpretability: unlike human-readable skill text, behaviors stored in LoRA adapters are more difficult to inspect and modify directly. Finally, the current setting assumes one skill per task and activates a single adapter during inference, whereas realistic agent workflows may require multiple interacting skills. Multi-skill composition, adapter routing, and behavioral conflict management therefore remain open problems for parameterized skill systems.

\FloatBarrier
\bibliography{main}

\end{document}